\documentclass[10pt,journal,compsoc]{IEEEtran}

\ifCLASSOPTIONcompsoc
  \usepackage[nocompress]{cite}
\else
  \usepackage{cite}
\fi

\ifCLASSINFOpdf

\else

\fi

\usepackage{epsfig}
\usepackage{graphicx}
\usepackage{amsmath}
\usepackage{amssymb}
\usepackage{comment}
\usepackage{multirow}
\newcommand{\Rmnum}{\mathrm{num}}

\usepackage{booktabs}
\usepackage{array, caption, threeparttable}
\usepackage{caption}
\usepackage{graphicx}
\usepackage{subfigure}
\usepackage[table,xcdraw]{xcolor}

\usepackage{algorithm}
\usepackage{listings}
\usepackage{color}

\usepackage{mathtools}

\usepackage{todonotes}
\usepackage{gensymb}
\usepackage{algorithm,algpseudocode}

\usepackage[breaklinks=true,bookmarks=false]{hyperref}

\DeclareUnicodeCharacter{2212}{-}

\begin{document}

\title{GM-DF: Generalized Multi-Scenario Deepfake Detection}

\author{Yingxin Lai, Zitong Yu, Jing Yang, Bin Li, Xiangui Kang, Linlin Shen

\thanks{Manuscript received May 2024. Corresponding author: Zitong Yu (email: zitong.yu@ieee.org). }

\thanks{This work was supported by Open Fund of National Engineering Laboratory for Big Data System Computing Technology (Grant No. SZU-BDSC-OF2024-02) and National Natural Science Foundation of China under Grant 62306061.}

\thanks{Y. Lai and J. Yang are with the School of Computing and Information Technology, Great Bay University, Dongguan 523000, China.}

\thanks{Z. Yu is with the School of Computing and Information Technology, Great Bay University, Dongguan 523000, China, and National Engineering Laboratory for Big Data System Computing Technology, Shenzhen University, Shenzhen 518060, China}

\thanks{B. Li is with the Guangdong Key Laboratory of Intelligent Information Processing, Shenzhen Key Laboratory of Media Security, Guangdong Laboratory of Artificial Intelligence and Digital Economy (SZ), Shenzhen Institute of Artificial Intelligence and Robotics for Society, Shenzhen University, Shenzhen 518060, China.}

\thanks{X. Kang is with the Guangdong Key Laboratory of Information Security, and the School of Computer Science and Engineering, Sun Yat-sen University, Guangzhou 510080, China}

\thanks{L. Shen is with Computer Vision Institute, School of Computer Science \& Software Engineering, Shenzhen Institute of
Artificial Intelligence and Robotics for Society, Guangdong Key Laboratory of
Intelligent Information Processing, and National Engineering Laboratory for Big Data System Computing Technology, Shenzhen University, Shenzhen 518060,
China. }

}

\markboth{IEEE Transactions on Dependable and Secure Computing}%
{Shell \MakeLowercase{\textit{et al.}}: Bare Advanced Demo of IEEEtran.cls for IEEE Computer Society Journals}

\IEEEtitleabstractindextext{%
\begin{abstract}

Existing face forgery detection usually follows the paradigm of training models in a single domain, which leads to limited generalization capacity when unseen scenarios and unknown attacks occur. In this paper, we elaborately investigate the generalization capacity of deepfake detection models when jointly trained on multiple face forgery detection datasets. We first find a rapid degradation of detection accuracy when models are directly trained on combined datasets due to the discrepancy across collection scenarios and generation methods. To address the above issue, a Generalized Multi-Scenario Deepfake Detection framework (GM-DF) is proposed to serve multiple real-world scenarios by a unified model. First, we propose a hybrid expert modeling approach for domain-specific real/forgery feature extraction. Besides, as for the commonality representation, we use CLIP to extract the common features for better aligning visual and textual features across domains. Meanwhile, we introduce a masked image reconstruction mechanism to force models to capture rich forged details. Finally, we supervise the models via a domain-aware meta-learning strategy to further enhance their generalization capacities. Specifically, we design a novel domain alignment loss to strongly align the distributions of the meta-test domains and meta-train domains. Thus, the updated models are able to represent both specific and common real/forgery features across multiple datasets. In consideration of the lack of study of multi-dataset training, we establish a new benchmark leveraging multi-source data to fairly evaluate the models' generalization capacity on unseen scenarios. Both qualitative and quantitative experiments on five datasets conducted on traditional protocols as well as the proposed benchmark demonstrate the effectiveness of our approach. The codes will be available on \href{https://github.com/laiyingxin2/GM-DF}{https://github.com/laiyingxin2/GM-DF}.


\end{abstract}

}

\maketitle

\IEEEdisplaynontitleabstractindextext

%
\IEEEpeerreviewmaketitle

\ifCLASSOPTIONcompsoc
\IEEEraisesectionheading{\section{Introduction}\label{sec:introduction}}
\else
\section{Introduction}
\label{sec:introduction}
\fi



\IEEEPARstart{A}{dvancements} in deep learning have facilitated the creation of face forgery mechanisms \cite{deepfake, fs, f2f, nt, yu2024benchmarking, shi2024shield}. These techniques simplify the generation of highly realistic forged face images, posing risks to both political and personal reputations and giving rise to significant social challenges. Consequently, the development of detection methods to mitigate these risks is imperative.
To alleviate discrepancies among various face forgery detection datasets, some researchers have adopted a specific approach. They treat the task of detecting forged faces as a binary classification task, utilizing existing deep convolutional neural networks to categorize the data into two distinct classes: real and forged. The primary goal of these investigations is to identify and extract common features to address the challenge of feature discrepancies. Several approaches have been proposed to tackle this issue, including the use of noise as a form of supervision \cite{capsule, srm}, the incorporation of frequency domain information \cite{freq1_icml, f3net, freq1_icml}, and the application of reconstruction techniques to gain insights into the distribution of authentic samples \cite{magdr, recce}.

\begin{figure}[t]
\centering
\label{tse}
\includegraphics[width=\linewidth]{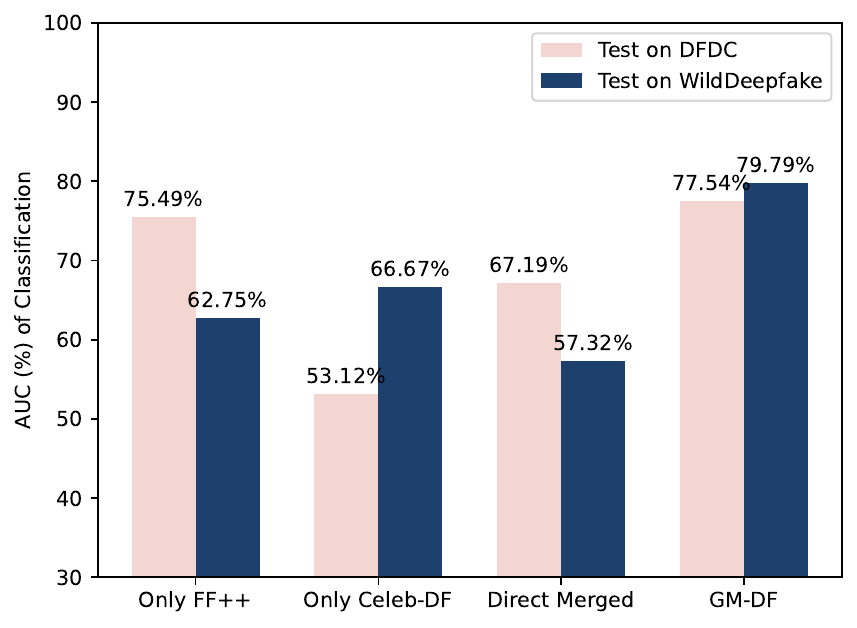}
\caption{Challenges in training a detector from multiple datasets. The generalization capacity of the baseline Xception \cite{xception} trained on FF++\cite{ff++}\&Celeb\cite{celebdf} datasets drops sharply while the proposed method GM-DF benefits obviously from multi-dataset training. }
\label{diffre_merge}
\end{figure}

However, despite the remarkable accuracy and precision attained by these models when applied in a cross-domain setting, their effectiveness remains heavily dependent upon the training process conducted on only one dataset. The initial strategy involves training a baseline model on the combined datasets. However, the results shown in Figure \ref{diffre_merge} indicate that direct training within combined datasets easily leads to generalization drops. The main reasons behind this might be the variances in forgery techniques, capturing circumstances, forgery methods, and hardware across various domains. In light of the continuous emergence of manipulated facial datasets, it is imperative to integrate and simultaneously train using different accessible data sources.
  
But if two face forgery detection datasets with different distributions are directly merged and used for training, the problem of domain conflict will inevitably be encountered. For example, as shown in Figure \ref{diffre_merge}, the merging of Celeb-DF(V2) \cite{celebdf} which the original FF++ \cite{ff++} dataset, suffers from degradation of accuracy from 75.49\% to 67.19\%. Therefore, the previous paradigm of single dataset training and testing does not work well on multiple domains, and the direct merging of individual datasets does not improve the generalization ability of the model well. With the increasing number of various face forgery detection datasets, \textit{how to effectively train a unified detector on multiple widely differentiated datasets is worth exploring. The solutions behind the problem might benefit the development of forgery foundation models.}

 Based on the above observations, in this paper, we propose a unified face forgery detection framework to solve the multi-dataset conflict problem, and our model is orthogonal to existing methods. We discover a novel insight: data conflict might be caused by ignoring the domain-specific features of the datasets. In order to enhance the models' generalization ability with the increasing number of datasets, we design a hybrid expert modeling approach to extract the domain-specific features while leveraging image-text alignment and masked image reconstruction mechanism to extract common real/forgery features across domains. Finally, 
 we supervise the models via a domain-aware meta-learning strategy. we design the novel domain alignment loss to strongly align the distributions of the meta-test domains and meta-train domains. Thus, the updated models are able to represent both specific and common real/forgery features across multiple datasets.

Extensive experiments are conducted on five public autopilot datasets, including FF++ \cite{ff++} Celeb-DF(V2) \cite{celebdf}, WildDeepfake \cite{wildfake}, DFDC \cite{dfdc} and the fake face dataset generated by diffusion DFF \cite{dff} to study the problem of data conflict in each domain or merged domains. Towards the era of large-scale multi-dataset training and testing, we establish a novel benchmark with five mainstream datasets, and the results show that the proposed models have strong generalization ability. Our contributions are as follows:
\begin{itemize}
    \item We are the first to comprehensively investigate the multi-dataset training task for face forgery detection, and establish a new benchmark leveraging multi-dataset data to fairly evaluate the models' generalization capacity.
    \item We propose a hybrid expert modeling approach for domain-specific real/forgery feature extraction. We also propose to represent common features via simultaneously aligning visual and textual features, and reconstructing masked faces across domains.
    \item We supervise the models via a domain-aware meta-learning strategy with a novel domain alignment loss.
    \item The proposed method achieve state-of-the-art performance on both traditional protocols as well as the proposed benchmark.
\end{itemize}

\vspace{-0.8em}
\section{Related Work}
\label{sec:relatedwork}

In this section, we briefly describe deepfake detection, vsion language models, and joint training on multiple datasets.

\subsection{Face Forgery Detection}
Recently, face forgery detection has received extensive attention from researchers due to the great threat to security and privacy. Previous methods \cite{xception,resnet,efficientnets,recurrent} treat face forgery detection as a binary classification problem in intra-dataset testing and has achieved very satisfactory performance. However, only relying on binary classification supervision, deepfake detectors easily overfit the training data. Thus, people began to turn to cross-domain performance exploration,  F$^3$-Net \cite{f3net} combines with the frequency domain information to extract the subtle differences between real and fake pictures, proved the effectiveness of the frequency domain in forgery detection artifact recognition. Similarly, SPSL \cite{spsl} proposes a frequency-based phase spectral analysis method. Face X-ray \cite{facexray} detects generated images by picture mixing boundaries, and DADF \cite{lai2023detect} leverages vision foundation model for robust forgery localization. PCL \cite{pcl} improves the supervisory performance by learning the inconsistency between the forged/neighboring regions and learning the commonality from real samples while reconstructing real samples. SLADD \cite{SLADD} combines data augmentation and face blending to improve the generalization ability.M-FAS \cite{kong2024m} and \cite{yu2024benchmarking} established a unified face forgery detection system.

Although these methods substantially improve the generalization ability, they are still limited by the common features and the specific forgery patterns in the training set, which aggravates the data Conflicts.

\begin{figure*}[h] 
\centering 
\label{tse}
\includegraphics[width=1.0\textwidth]{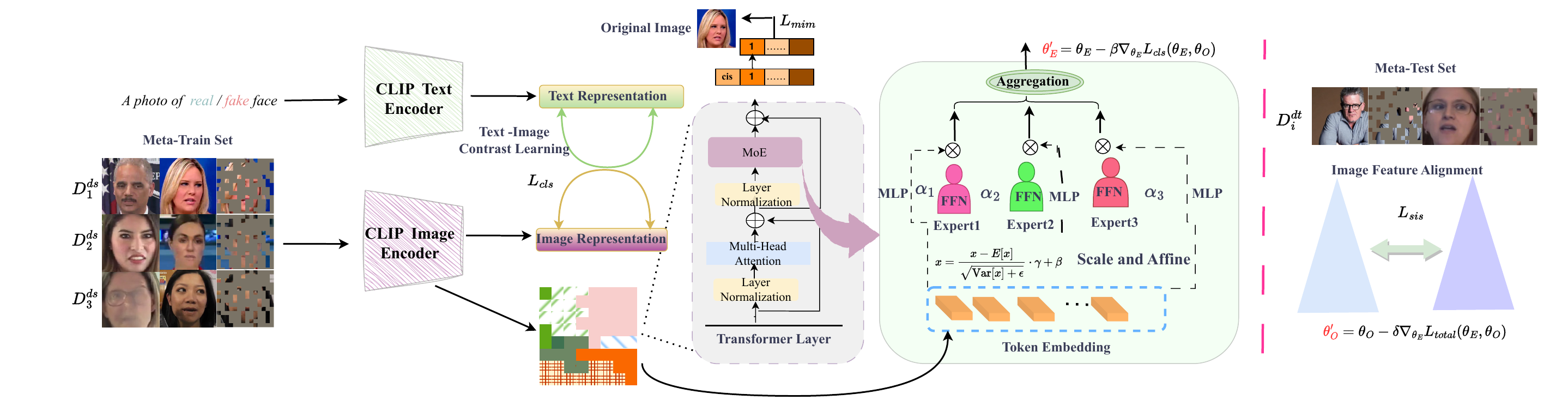}
\caption{The framework of the proposed method. It integrates meta-learning modeling with image-text contrastive learning. It comprises three pivotal components: Dataset-Embedding Generator (DEG) and a Multi-Dataset Representation (MDP), as well as a Meta-Domain-Embedding Optimizer(MDEO). Firstly, the DEG incorporates a Dataset Information Layer (DIL) and a dynamic text feature affine aimed at mapping discriminative features unique to each domain, and the second part MDP is the face mask image modeling (MIM) reconstruction module, which provides additional detail information for the global features of CLIP. To consider the difference between each domain, we propose to use the higher-order statistical features in Domain Alignment (DA) loss to constrain the feature distribution. In this process, MDEO was used to optimize the learned two features.
} 
\label{diffre} 
\end{figure*}

\vspace{-1.0em}
 \subsection{Vision Language Models}
Visual-language models are rich in multimodal feature representations and show surprising generalization performance in downstream tasks. \cite{tip_adapter} proposes an adaptive approach to CLIP \cite{clip} modules without training that performs state-of-the-art small-sample classification tasks on ImageNet. CoOp \cite{coop} aims to introduce a learnable Prompt approach to better adapt powerful and generalized a priori of visual-linguistic models to downstream tasks. OpenCLIP \cite{openclip} The model integrates the cross-modal capabilities of text encoder with the generative abilities of the pre-trained language model BART, resulting in a strengthened text encoder for language bachbone.Lit \cite{lit} utilizes multimodal pre-trained models to improve graphic alignment. Flamingo \cite{flamingo} predicts the next text token based on the previous text and the visual Token, thus better introducing visual information for text creation.  
LLAVA \cite{LLAVA} proposes a command optimization technique for vision. BLIP2 \cite{blip2} designs Q-Former to bridge between visual and linguistic models by connecting temporal and linguistic features. Although these methods achieve good generalization performance in downstream tasks, they face the challenges of high computational effort and complexity. In addition, they are mostly applied to face forgery detection, where lack of robustness remains a problem.


\vspace{-1.0em}

\subsection{Joint Training on Multiple Datasets}
 For traditional image tasks such as target detection 
\cite{fastrcnn,yolov4} and semantic segmentation \cite{unet,swinunet}, due to the different dataset class labels and fine-grained cross-dataset difference, they result in poor generalization when directly fusied data for training. Some researchers have begun to study the data federation \cite{dynamic, mdalu,towards,simple,mseg}. Dai et al. \cite{dynamic} combine multiple self-attention mechanisms sequentially to unify the target detection head \cite{mseg} and relabeling the instances of disaggregation to perform the alignment operation on to the images significantly improve the generalization ability of the model. Wang et al. \cite{ towards} trains a generalized object detector by incorporating different supervised signals, eliminating the need to model differences across data. Zhao et al. \cite{object} propose a pseudo-labeling method that is tuned for specific situations, showing that a unified detector trained on multiple datasets can outperform each detector trained on a specific dataset.
 Although recent works on the generic image classification task using multi-domain data for training have been partially investigated, it has not been explored in the field of face forgery detection. Moreover, different training domains are not equally important due to variant environments, media quality, and attack types.  Such biased and imbalanced data from different domains makes this task challenging.

\section{Methodology}

The framework of the proposed GM-DF is shown in Figure \ref{diffre}, which contains a Dataset-Embedding Generator (DEG) and a Multi-dataset Representation (MDP), as well as a Meta-Domain-Embendding Optimizer (MDEO). The DEG pay attention to information that is unique to the dataset,the MDP focuses on learning more fine-grained, local relational features of forged patterns, whereas the MDEO achieves its functionality by modeling the relationships between universal information and dataset embending. For better understanding, we provide some brief details before outlining the framework architecture.  
 
\subsection{Preliminary}
To solve the problem of poor cross-domain performance for multiple scenarios and datasets, we first assume that this is due to domain differences. As shown in Table \ref{datasets}, different datasets have different collection scenarios and forgery methods. Currently, there is a trend towards diversification in sources of data for facial forgery detection. Figure \ref{difface} highlights distinct differences among various datasets. For instance, the DFDC \cite{dfdc} dataset exhibits a prevalence of green backgrounds, WDF \cite{wildfake} images convey an impression of magnification, DFF \cite{dff} tends towards an artistic mode of photography, while the forgeries in Celeb-DF appear relatively homogeneous, and FF++ \cite{ff++} features facial representations with rich attributes. Mixing these datasets may lead to model learning biases due to their inherent disparities.	   

\begin{figure}[t]
\centering  
\subfigure[DFF]{%
\includegraphics[width=0.3\linewidth]{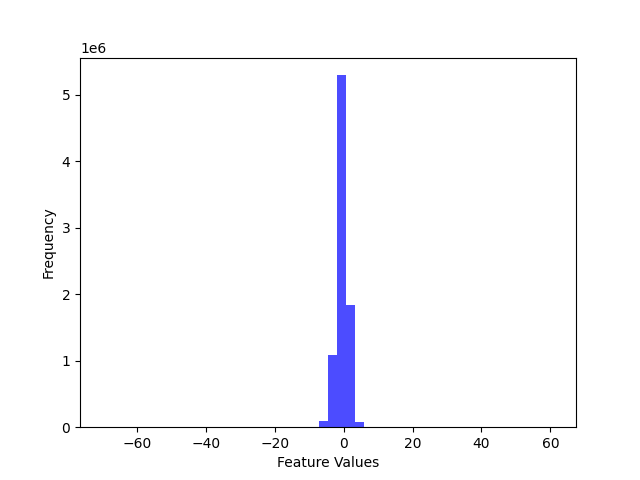}}
\subfigure[FF++]{%
\includegraphics[width=0.3\linewidth]{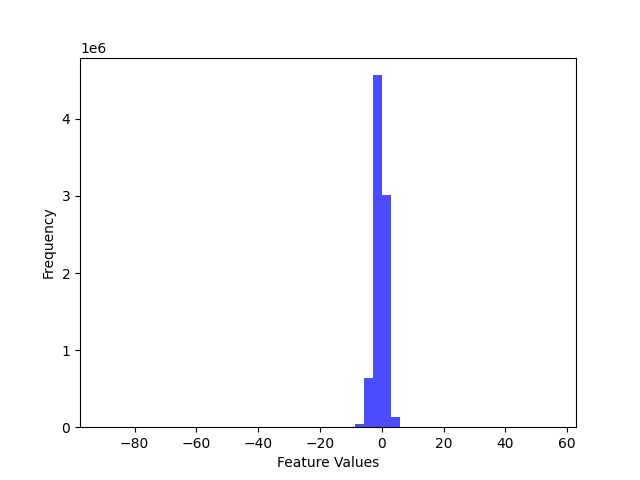}}\\
\subfigure[Celeb-DF]{%
\includegraphics[width=0.3\linewidth]{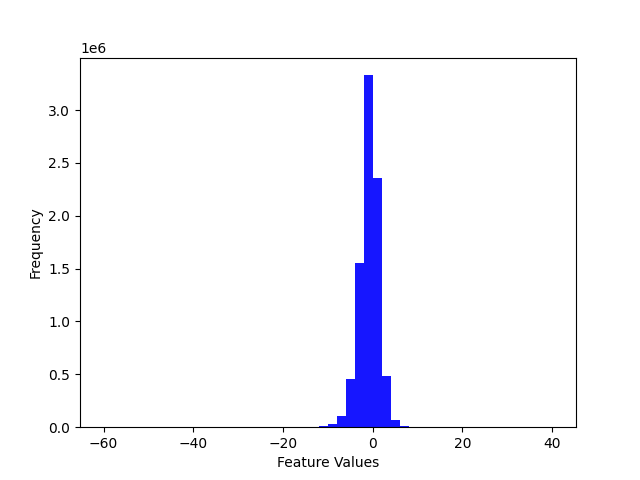}}
\subfigure[WDF]{%
\includegraphics[width=0.3\linewidth]{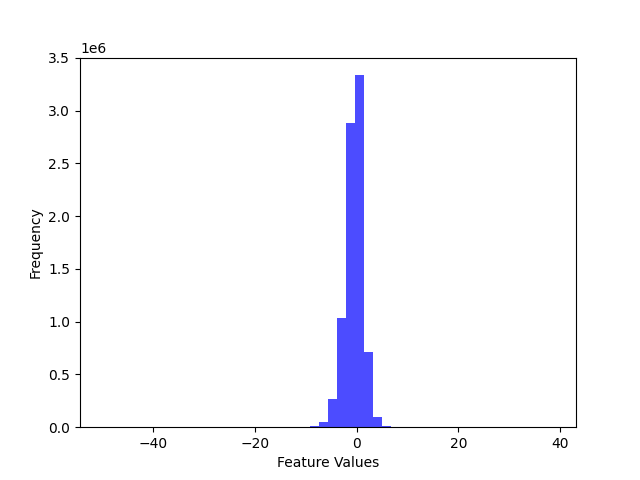}}
\caption{Histograms of feature values in a randomly selected channel, where features are computed from the block of a convolution based on Xception \cite{xception} trained on the dataset of four domains \cite{wildfake,dff,ff++,celebdf}.}
\label{distribution}
\end{figure}

\begin{figure}[t] 
\centering 
\label{Datavis}
\includegraphics[width=0.5\textwidth]{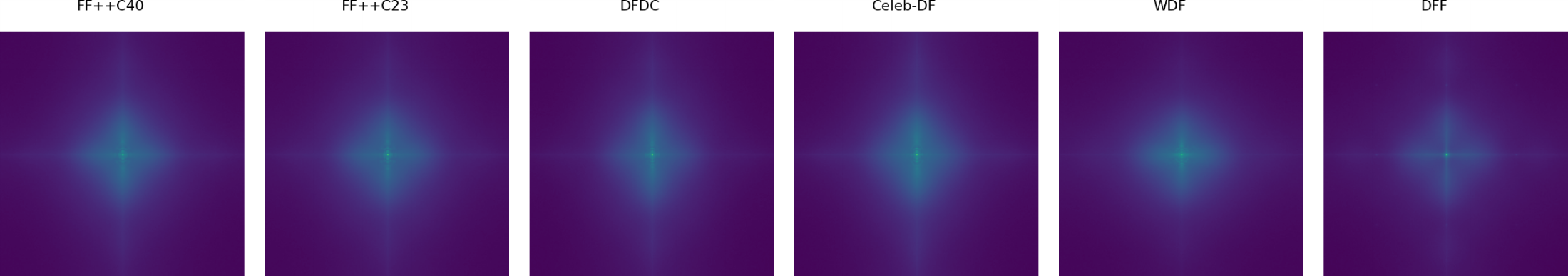}
\caption{
The commonly used frequency domain detection model M2TR's \cite{M2TR} frequency domain visualization on the FF++ c40 \cite{ff++}, FF++ c23 \cite{ff++}, DFDC \cite{dfdc}, Celeb-DF \cite{celebdf}, WildDeepfake \cite{wildfake}, and DFF \cite{dff} datasets.} 
\label{frequency} 
\end{figure}

We also observe that current deepfake detectors \cite{xception,recce,f3net,ucf} usually focus on representing common patterns. As shown in Figure \ref{distribution}, the distribution of feature differences in each dataset is relatively small, indicating that the model learned some common features but ignored the specific features of each domain. The utilization of frequency domain information for detection is a widely employed technique. As depicted in Figure \ref{frequency}, these methods merely capture singular counterfeit and learning patterns. The consistent frequency domain visualizations across various datasets underscore the imperative nature of learning dataset characteristics.

It is also worth noting that the DFF \cite{dff} dataset generated by diffusion method is also not very different from the other datasets, leading to large differences in their domains and thus the cause of domain conflicts, so we would like to set up a more specific model that reduces domain conflicts, learns more characteristic forgery features after mapping to the feature space, and at the same time can be well generalized to catch the differences between real and fake images, since forgery patterns are usually hidden in low-level details. Therefore, we refer to the principle of Adaptive Risk Minimization \cite{adaptive}, which aims at co-optimal solutions in multiple domains. Specifically,
here, we describe our adaptive modelling. Divided into $N$ source domains $D=\{d_{s^{1}},d_{s^{2}}\cdots d_{s^{n}} \}$ represent source face forgery datasets and $M$ target domains. Define $D_{t}=\{d_{t^{1}},d_{t^{2}} \}$ where each domain has input and label. Using $x\in X$ and $y\in Y$ as input and label, we may define the source domains as $D_{s}=\{x_{s^{i}},y_{s^{i}} \}$. To simulate real-world cross-domain challenges by mimicking test-time adaptation (i.e., adjusting prior to prediction), we use characteristic domain weights in the inner loop to learn information unique to each domain, and reconstruction learning and distributional approximation in the outer loop to allow the model to learn the differences between real and fake images.

\subsection{Dataset--Embedding Generator}

After training use the underlying source domain dataset, it is usually possible to extract a large number of visual features that match the characteristics of the domain. However, when confronted with unseen scenarios and unknown forgery categories, models (e.g., Xception) usually have poor feature generalization (see Figure \ref{diffre_merge}). This is mainly due to the significant semantic differences between the forgery patterns of the new category and those in the underlying dataset.

For example, when a model processes an image of a face collected under a curtain, it may incorrectly misinterpret features such as the eyes and nose of the face as features of the forgery image under the curtain. This is because there may be some false pictures under the curtain in the underlying dataset, leading to confusion in the model's learning process. This situation prevents the model from correctly recognizing the forgery images in certain environments or situations. To mitigate this problem, we explore additional semantic information cues to guide the visual feature network to obtain rich and flexible semantic features.

Specifically we use Vit as the foundation model for fine-tuning due to its unbiasedness for each category of both real and fake images and language modeling's potential.
This module follows the Mixture of Experts (MoE) \cite{masoudnia2014mixture} network structure to build a mixture of expert layers to learn domain-invariant features; unlike the domain-specific module we propose based on this, we use $N$ independent experts. Each residual block consists of a Dataset Information Layer and an Multilayer Perceptron (MLP), 
Since the domain-specific embendding is much smaller than the normal backbone, it can be used if there is a low additional computational cost and restrain the trends of the overfit,
experts from various domains carried out the process to extract domain-invariant and domain-specific features as follows:
\begin{equation}
F_(x) = F_\theta(x) + \Delta F^{n}_{\theta}(x)
\end{equation}

Here, $F_{\theta}(x)$ represents the original function that is shared by all source domains to learn the common domain-invariant features. $ \Delta F^{n}_{\theta}$ adaptively extracts the discriminative and unique domain-specific features.

Although existing works show that the activations in different transformer blocks contribute to the stability of the training, the diversity of the individual domains is sacrificed in the case of multi-domain training. So we model each expert in MoE layer via introducing a new Dataset Information Layer (DIL) with domain-specific parameters.
Unlike the fixed gain and bias  in LayerNorm, we add skip connections and then scale the function by a learnable parameter called the domain weights and initialize it to 0 at the outset. The signal propagates as follows:

\begin{equation}
\begin{gathered}
x_{i+1}=\alpha_i*\text{Sublayer}(x_i),\\
\end{gathered}
\end{equation}
where $\text{Sublayer} \subseteq \left\{\text{self-attention}, \text{feed-forward} \right\}$,
Then, we compute the gain and bias with respect to the learned prompt
vector $w$.
where $\alpha_i$
is the learned residual domain weight parameter. At the initial time, all the $\alpha_i$ are initial to zero;  the network represents a constant function, at which point dynamic equidistance is directly satisfied. Then the model gradually learns the specific features corresponding to each domain.
In order to allow models to learn their respective domain-specific knowledge through the parameters and to dynamically generate them in real time according to different instances, we use the learned  prompt vector to perform an affine to the normalized input features based on VPT \cite{visualprompt}.
More precisely, given domain $d_{t}^{i}$ and prompt feature vector
$p=\left[ v_{1} ,v_{2},v_{3}\cdots ,v_{M}\right]$ where $M$ is the dimensionality of the learnable prompt vector, we derive a MLP layer $ h\left(\cdot \right)$ to the specific feature


\begin{equation}
\text{DIL}\left( v,p \right) = h\left( p \right) \cdot x_{i+1} 
\end{equation}

The entire transformer block can be formalized as follows:
\begin{equation}
\begin{split}
x_0 = \text{LayerNormal}_{\text{att}_i}(x),x = \text{MHA}(x_0) + x, \\ 
x'_{0} = \text{DIL}_{\text{moe}_i}(x),x = \text{MoE}(x'_{0}) + x.
\end{split}
\end{equation}

The input features first go through the original transformer layernorm $\text{LayerNormal}_{\text{att}_i}$ as well as the multi-head attention $\text{MHA}$ , and then through the various expert modules $\text{DIL}_{\text{moe}_i}$ and $\text{MoE}$.




\subsection{Multi-Dataset Representation}

After obtaining the domain embedding and expert views, we calculate
the scaled dot-product attention and mark it as the expert
views, which is formulated as [11]:
\begin{equation}
\text{Attention}(Q, K, V) = \text{softmax}\left(\frac{QK^T}{\sqrt{d_k}}\right) V,
\end{equation}
where $Q$ denotes the query, $K$ denotes the key, $\sum_{}^{}$ denotes the value of the input
embedding, and the scale factor of $d_k$ is the key of dimension. Here
we compute the attention score containing the task information
Setting $Q$ and $K$ as the
\begin{equation}
Q = K =  \text{Concat}(\Delta F_{\theta_1}(x), \Delta F_{\theta_2}(x), \ldots, \Delta F_{\theta_N}(x)) \in \mathbb{R}^{1 \times N},
\end{equation}
where Concat denotes the operation that stacks vectors into a matrix. $V$ is a matrix stacked by expert views. We make the
summation of the expert views to obtain the task-specific aggregated expert view.

Image-text pairs can learn semantic feature representations of face forgeries about specifics, but they may not be able to capture the details. Inspired by previous study on forgery face reconstruction properties \cite{recce,ucf,yang2023masked} and to improve face detail representation, we add a mask image modeling (MIM) \cite{he2022masked} task that masks a number of patches of the input image and predicts their visual tokens. Commonly used, typical low-level visual tasks mask the image to capture low-level details and offer semantic information.With the learned representations, the reconstruction difference of real and fake faces significantly differs in distribution.

Given an input image $X$, we begin by dividing it into $N$ patches denoted as $\{x_1, x_2, x_3, \ldots, x_n\}$, where $n$ represents the total number of patches. Subsequently, we adopt a stochastic masking approach, referred to as \cite{tian2020makes} to apply masks to a subset of $M$ patches. This process results in a modified image $X'$, expressed as $X' = \{x_1, x_{m2}', x_{m3}', \ldots, x_n\}$. Here, $x_{m2}'$ means that the second one is replaced by a mask.
Next, we feed the masked images into a shared Transformer architecture, yielding a set of hidden vectors $\{h'_{\text{cls}}, h'_1, h'_2, \ldots, h'_N\}$. Leveraging the knowledge encapsulated in these hidden vectors, we proceed to predict the masked regions $\{x_{m_i}' \mid m_i \in M\}$ and simultaneously perform direct pixel-level predictions.

To optimize memory consumption, a Gumbel-Softmax Variational Autoencoder \cite{categorical} is employed. Each image block is encoded into one of $T$ possible values, and a classification layer operates within the hidden vector space to indirectly predict the indices of the masks.
The loss function is given as:
\begin{equation}
\mathcal{L}_{\text{mim}} = - \sum_{k \in M} \log p\left(q_k^\phi(x) | x^{'}\right).
\end{equation}


Here, $p(q_k^\phi(x) | \tilde{x})$ represents the classification score for classifying the $k$-th hidden vector belonging to the visual token $q_k^\phi(x)$, where $q_\phi$ is a categorical distribution.


Due to domain discrepancy, it is difficult to let models learn the intrinsic differences of different domains by themselves. How to mine the key universal information across domains to feedback to the model? To this end, we design the Domain Alignment (DA) loss of each domain and meta-test domain based on the distribution to align the distribution to a specific domain. First, the eigenmeans of the training set are $\mu_{\text{source}}$, the covariance matrix is $\Sigma_{\text{s}}$, the eigenmeans of the generated samples are $\mu_{\text{s}}$,the covariance matrix is $\Sigma_{\text{t}}$.
\begin{equation}
\mathcal{L}_{\text{sis}} = \|\mu_{\text{s}} - \mu_{\text{t}}\|^2 + \text{Tr}(\Sigma_{\text{s}} + \Sigma_{\text{t}} - 2(\Sigma_{\text{s}}\Sigma_{\text{t}})^{1/2}).
\end{equation}

Based on the feature prior, it is instantiated as they calculate the distance between two distributions with mean and covariance matrices.Smaller distances represent that source domains is closer to the target domain distribution.







\begin{algorithm}[t]\small
\small
\caption{Training for Meta Deepfake Detection}
\begin{algorithmic}
\State \textbf{Input} $D$: data of multi-source domains; $\delta, \beta$: learning rates; 

\State \textbf{Initialize:} $\theta_E, \theta_O$
\While{not converged}
    \State Sample $N-1$ domains as meta-train set $D^{ds}_{i}$ and
the remaining domain as meta-test set $D^{dt}_{i}$.
    \For{each $D^{ds}_{i}$}
        \State Evaluate loss $L_{cls}$ on $D^{ds}_{i}$ 
        \State Update $\theta_{E}$ by: $\theta_{E}^{\prime}\gets 
        \theta_{E}-\beta\frac{\sigma L_{cls}\left( f\left( \theta_{E}, \theta_{O}\right) \right)}{\sigma\theta_{E}}$
    \EndFor
    
    \State Update$A$  $\theta_E, \theta_O$ for the current meta batch:
    \State Evaluate loss $L_{sis}$  and $L_{mim}$ on $D^{dt}_{i}$ 
    \State Update $\theta_{O}$ by: $\theta_{O}^{\prime}\gets \theta_{O}-\delta\frac{\sigma L_{total}\left( f\left( \theta_{E}, \theta_{O}\right) \right)}{\sigma\theta_{O}}$
\EndWhile
\end{algorithmic}
\label{algorithm}
\end{algorithm}

\subsection{Meta-Domain-Embedding Optimizer}
In this subsection, we propose a meta-domain-embedding optimizer based on the MAML \cite{finn2017model} paradigm (see Algorithm \ref{algorithm}) for pouncing on the generic and personality feature capabilities of learning domain-specific and domain-common features. Here we define each domain as a single task $t$. 
In the training process we sample batches of multi-domain data, which consist of meta-train set $D^{ds}_{i}$ and meta-test set $D^{dt}_{i}$, here for simplicity we assume that the full model is described as a function $f\left( \cdot  \right)$, which receives an image $x$ as input and $y$ as output. The loss function optimized per meta-train domain task during the training is uses cross-entropy loss defined as
\begin{align}
L_{cls}\left( {f\left( \theta_{E},\theta_{O}\right)} \right) &= \sum_{(x_j, y_j) \in D_{d_i}} \left[ y_j \log f(x_j) \right. \nonumber \\
&\quad \left. + (1 - y_j) \log(1 - f(x_j)) \right].
\end{align}
   
In this process referred as the inner-loop update, importantly, we just update the learnable token parameter in the meta train and freeze all other feature extraction. $\theta_{E}$ represents the meta-MoE's expert and vpt parameters, while $\theta_{O}$ represents the base model's parameters. After generating the initial domain embeddings $\theta_{i}$  and evaluating the obtained losses on the batch of data, obtains the updated domain embeddings by calculating the gradient of the losses $L_{cls}$ and performing gradient descent updates.
\begin{align}
\theta_{E}^{'}\gets \theta_{E}-\beta\frac{\sigma L_{cls}\left( f\left( \theta_{E}, \theta_{O}\right) \right)}{\sigma\theta_{E}},
\end{align}
where $\beta$ is the learning rate of gradient descent.
In the subsequent step, the model's meta-parameters $\theta_{E}^{'}$ undergo optimization to enhance the performance of meta-test set $D^{ds}_{i}$ to get the loss  $L_{cls}$ and the prediction for domain $i$.

Similarly, during the meta-test phase, the meta-test sample  $D^{dt}_{i}$is utilized to update the network. The features are aggregated using the aggregation model after passing through the expert layer. Additionally, the consistency loss $\text{L}_{\text{sis}}$ of the features is employed to minimize the distance between the source domain and the target domain with reconstructed facial features aid fine-grained forgery feature learning.
The overall model loss is stated as follows
\begin{align}
L_{total}=L_{sis}+L_{cls}+L_{mim}.
\end{align}
Then we can optimize the generator $f\left( \cdot  \right)$  by the gradient:
\begin{align}
 \theta_{O}^{'}\gets \theta_{O}-\delta\frac{\sigma L_{total}\left( f\left( \theta_{E}, \theta_{O}\right) \right)}{\sigma\theta_{O}}.
\end{align}

In summary, $\theta_{E}$ is updated during the meta-train process to learn the private characteristics of each domain and has higher flexibility due to the dynamic prompt vector. $\theta_{O}$ is updated during the meta-test process to capture generic forged clues, which helps the model acquire complementary information and be used for multi-domain training.

\begin{table}[t]
  \caption{Information of the datasets used in our protocols}
  \label{datasets}
  \resizebox{0.47\textwidth}{!}{
  \begin{tabular}{|c|c|c|c|}
      \hline
      Source Dataset       & Collected from & Synthesis Methods                                                                                    & Identity \\ \hline
      Faceforensics++ \cite{ff++}     & YouTube        & \begin{tabular}[c]{@{}c@{}}DeepFake/Face2Face/\\ FaceSwap/NeuralTextures\end{tabular}                & -        \\ \hline
      Celeb-DF (V2)  \cite{celebdf}        & YouTube        & Improved Deepfake                                                                                   & 59+      \\ \hline
      DFDC    \cite{dfdc}             & Actors         & \begin{tabular}[c]{@{}c@{}}StyleGAN\\ FSGAN\\ Refinement\\ Audioswaps\\ NTH\end{tabular}             & 960      \\ \hline
      Deepfake in the Wild \cite{wildfake} & Internet       & Unknown                                                                                             & 100      \\ \hline
      DeepFakeFace    \cite{dff}                & IMDB/Wikipedia & \begin{tabular}[c]{@{}c@{}} Stable Diffusion/Inpainting/Insight\end{tabular} & -        \\ \hline
    \end{tabular}
    }
\end{table}

\begin{table*}[]
\caption{Results (AUC (\%) and ACC (\%)) of joint training on FF++ \cite{ff++}, Celeb-DF \cite{celebdf} , and DFF \cite{dff} datasets. }
\label{compare2}
\scriptsize
\centering
\label{compare2}
\begin{tabular}{cccccccccccc}
\hline
 &  & \multicolumn{4}{c}{Cross-Domain} & \multicolumn{6}{c}{In-Domain} \\ \cline{3-12} 
 &  & \multicolumn{2}{c}{Tested on DFDC \cite{dfdc}} & \multicolumn{2}{c}{Tested on WDF \cite{wildfake}} & \multicolumn{2}{c}{Test On FF++ \cite{ff++}} & \multicolumn{2}{c}{Test On Celeb \cite{celebdf}} & \multicolumn{2}{c}{Test On DFF \cite{dff}} \\ \cline{3-12} 
\multirow{-3}{*}{Source Domain} & \multirow{-3}{*}{Baseline Method} & AUC(\%) & \multicolumn{1}{c}{ACC(\%)} & AUC(\%) & ACC(\%) & AUC(\%) & \multicolumn{1}{c}{ACC(\%)} & AUC(\%) & \multicolumn{1}{c}{ACC(\%)} & AUC(\%) & ACC(\%) \\ \hline
 &  Xception \cite{xception} $\left( ICCV \,2019\right)$ & 75.49 & \multicolumn{1}{c}{53.82} & 62.74 & 57.36 & 100 & \multicolumn{1}{c}{97.87} & 89.10 & \multicolumn{1}{c}{90.24} & 89.73 & 93.12 \\
 
 & REECE \cite{recce} $\left(CVPR \,2022\right)$ & {\color[HTML]{343434} 75.19} & \multicolumn{1}{c}{{\color[HTML]{343434} 73.42}} & {\color[HTML]{343434} 77.90} & 62.18 & 99.98 & \multicolumn{1}{c}{98.17} & 92.31 & \multicolumn{1}{c}{94.16} & 93.11 & 94.38 \\
  
 & UCF \cite{ucf} $\left( ICCV \,2023\right)$ & {\color[HTML]{343434} 80.50} & \multicolumn{1}{c}{{\color[HTML]{343434} 73.01}} & {\color[HTML]{343434} 73.40} & 67.52 & 98.72 & \multicolumn{1}{c}{99.60} & 82.40 & \multicolumn{1}{c}{86.14} & 93.11 & 94.38 \\

  & Implicit \cite{recce}  $\left( CVPR \,2023\right)$& {\color[HTML]{343434} 74.90} & \multicolumn{1}{c}{{\color[HTML]{343434} 72.13}} & {\color[HTML]{343434} 75.12} & 69.40 & 99.98 & \multicolumn{1}{c}{96.23} & 82.80 & \multicolumn{1}{c}{83.19} & 90.11 & 92.80 \\

\multirow{-3}{*}{FF++ \cite{ff++}} & CLIP \cite{clip} $\left( ICML \,2021\right)$ & 76.01 & \multicolumn{1}{c}{72.51} & 74.33 & 64.52 & 93.21 & \multicolumn{1}{c}{96.10} & 81.43 & \multicolumn{1}{c}{83.71} & 92.21 & 93.19 \\
\hline 
 
 & Xception \cite{xception}  $\left( ICCV \,2019\right)$ & 53.12 & \multicolumn{1}{c}{52.16} & 66.67 & 43.75 & 51.32 & \multicolumn{1}{c}{54.62} & 96.32 & \multicolumn{1}{c}{98.61} & 73.10 & 75.46 \\
 
 & REECE  \cite{recce} $\left( CVPR \,2022\right)$ & 57.26 & \multicolumn{1}{c}{54.71} & 69.32 & 67.15 & 53.17 & \multicolumn{1}{c}{55.71} & 99.20 & \multicolumn{1}{c}{99.33} & 76.32 & 74.53 \\
 
 & UCF  \cite{ucf} $\left( ICCV \,2023\right)$& 65.04 & \multicolumn{1}{c}{62.90} & 61.24 & 63.90 & 61.46 & \multicolumn{1}{c}{63.09} & 97.12 & \multicolumn{1}{c}{97.06} & 73.71 & 72.33 \\

  & Implicit  \cite{implicit} $\left( CVPR \,2023\right)$& 64.60 & \multicolumn{1}{c}{62.12} & 65.11 & 64.54 & 61.08 & \multicolumn{1}{c}{65.71} & 99.20 & \multicolumn{1}{c}{93.33} & 76.32 & 71.02 \\

\multirow{-3}{*}{Celeb \cite{celebdf}} & CLIP \cite{clip} $\left( ICML \,2021\right)$& 54.32 & \multicolumn{1}{c}{51.67} & 65.17 & 61.34 & 51.03 & \multicolumn{1}{c}{52.00} & 96.98 & \multicolumn{1}{c}{93.12} & 72.33 & 76.45 \\ 
\hline  
  
 & Xception \cite{xception} $\left( ICCV \,2019\right)$& 67.19 & \multicolumn{1}{c}{56.62} & 59.31 & 56.22 & 82.99 & \multicolumn{1}{c}{68.75} & 100 & \multicolumn{1}{c}{96.87} & 93.89 & 91.23 \\
 
 & REECE \cite{recce} $\left( CVPR \,2022\right)$& 70.32 & \multicolumn{1}{c}{63.18} & 64.61 & 62.83 & 85.24 & \multicolumn{1}{c}{73.75} & 100 & \multicolumn{1}{c}{98.25} & 94.21 & 93.54 \\

 & UCF  \cite{ucf} $\left( ICCV \,2023\right)$& 65.43 & \multicolumn{1}{c}{63.21} & 67.50 & 63.96 & 85.72 & \multicolumn{1}{c}{82.10} & 97.98 & \multicolumn{1}{c}{97.04} & 89.32 & 87.49 \\
 
  & Implicit  \cite{implicit} $\left( CVPR \,2023\right)$& 57.26 & \multicolumn{1}{c}{67.81} & 63.50 & 81.70 & 78.50 & \multicolumn{1}{c}{75.41} & 98.41 & \multicolumn{1}{c}{98.10} & 89.63 & 86.10 \\

 & CLIP \cite{clip}  $\left( ICML \,2021\right)$& 67.41 & \multicolumn{1}{c}{62.78} & 65.78 & 63.08 & 83.51 & \multicolumn{1}{c}{76.10} & 100 & \multicolumn{1}{c}{97.71} & 93.53 & 92.15 \\

 \hline
 \multirow{-6}{*}{\shortstack{FF++  \cite{ff++}\\ \& Celeb \cite{celebdf}}}&

 Xception \cite{xception} $\left( ICCV \,2019\right)$& 58.82 & \multicolumn{1}{c}{53.12} & 58.82 & 53.12 & 96.13 & \multicolumn{1}{c}{{\color[HTML]{343434} 89.26}} & 99.79 & \multicolumn{1}{c}{96.55} & {\color[HTML]{343434} 93.93} & 95.41 \\

 & REECE \cite{recce} $\left( CVPR \,2022\right)$& 73.16 & \multicolumn{1}{c}{67.09} & 66.04 & {\color[HTML]{343434} 63.00} & {\color[HTML]{343434} 97.12} & \multicolumn{1}{c}{{\color[HTML]{000000} 91.19}} & {\color[HTML]{343434} 99.38} & \multicolumn{1}{c}{{\color[HTML]{343434} 98.19}} & 95.88 & {\color[HTML]{343434} 96.17} \\
 
 & UCF  \cite{ucf} $\left( ICCV \,2023\right)$& 67.40 & \multicolumn{1}{c}{59.70} & 72.10 & 69.54 & 85.29 & \multicolumn{1}{c}{83.50} & 98.74 & \multicolumn{1}{c}{97.41} & 92.32 & 92.08 \\

  & Implicit  \cite{implicit} $\left( CVPR \,2023\right)$& 68.91 & \multicolumn{1}{c}{61.43} & 69.32 & 69.54 & 89.32 & \multicolumn{1}{c}{89.40} & 99.20 & \multicolumn{1}{c}{99.33} & 90.04 & 90.56 \\

\multirow{-4}{*}{\shortstack{FF++ \cite{ff++}\\  \&Celeb \cite{celebdf}\\ \& DFF \cite{dff}}}
& CLIP \cite{clip} $\left( ICML \,2021\right)$& 55.43 & \multicolumn{1}{c}{51.02} & 73.45 & 71.07 & 96.26 & \multicolumn{1}{c}{85.95} & 99.24 & \multicolumn{1}{c}{97.23} & 94.71 & 93.23 \\ 

   &\textbf{GM-DF (Ours)}& \textbf{77.54}  & \multicolumn{1}{c}{\textbf{75.23}} & \textbf{79.70} & \textbf{75.08}& \textbf{98.23} & \multicolumn{1}{c}{\textbf{97.23}} & \textbf{99.99} & \multicolumn{1}{c}{\textbf{98.45}} & \textbf{97.72} & \textbf{98.78} \\ \hline
\end{tabular}


\end{table*}

\begin{table*}[t]
\caption{The results on the Multi-Domain Deepfake detection benchmarks based on $M_{EER}$(\%) and AUC (\%).}
\centering
\label{protocol}
\resizebox{0.8\textwidth}{!}{%
\begin{tabular}{lcccccccc}
\hline
\multirow{2}{*}{Method} & \multicolumn{2}{c}{FF++\& Celeb\&DFF} & \multicolumn{2}{c}{FF++\&Celeb\&WDF} & \multicolumn{2}{c}{FF++\&DFF\&WDF} & \multicolumn{2}{c}{Celeb\& DFF\&WDF} \\ \cline{2-9}
 & $M_{EER}$(\%) & AUC(\%) & $M_{EER}$(\%) & AUC(\%) & $M_{EER}$(\%) & AUC(\%) & $M_{EER}$(\%) & AUC(\%) \\ \hline
MesoNet  \cite{mesonet} $\left( WIFS \,2018\right)$ & 45.31 & 53.34 & 44.70 & 69.25 & 46.42 & 57.16 & 40.54 & 54.92 \\

Multl-task \cite{multitask} $\left( BTAS \,2019\right)$& 36.33 & 57.33 & 36.98 & 66.03 & 39.14 & 74.72 & 34.91 & 62.17 \\
F$^3$-Net \cite{f3net}  $\left( ECCV \,2020\right)$& 36.12 & 57.76 & 35.82 & 68.35 & 37.23 & 67.53 & 31.05 & 66.89 \\
Xception \cite{xception} $\left( ICCV \,2021\right)$ & 33.45 & 71.09 & 37.68 & 66.64 & 35.56 & 76.88 & 33.30 & 65.40 \\
REECE \cite{recce} $\left( CVPR \,2022\right)$& 32.57 & 70.85 & 35.07 & 69.86 & 36.64 & 78.72 & 30.14 & 71.92 \\
UCF \cite{ucf} $\left( ICCV \,2023\right)$ & 35.90 & 69.72 & 34.78 & 65.41 & 35.02 & 74.13 & 34.02 & 69.11 \\
Implicit \cite{implicit} $\left( CVPR \,2023\right)$ & 33.09 & 69.54 & 37.66 & 72.12 & 38.91 & 74.10 & 33.18 & 68.31 \\

\textbf{GM-DF (Ours)} & \textbf{30.13} & \textbf{74.33} & \textbf{32.81} & \textbf{72.37} & 
\textbf{32.90}  & \textbf{80.19}  & \textbf{28.36} & \textbf{73.73} \\ \hline
\end{tabular}%
}

\end{table*}

\section{Multi-Domain Deepfake Detection Protocols}
\label{protocls}

Towards the era of large-scale multi-dataset training and cross-dataset testing, we establish a novel benchmark with five mainstream datasets, including FaceForensics++ (FF++) \cite{ff++}, Celeb-DF (v2) (Celeb for short) \cite{celebdf}, WildDeepfake (WDF) \cite{wildfake}, and DFDC \cite{dfdc}. Information and visualization of these datasets can be found in Table \ref{datasets} and Figure \ref{difface}, respectively. 

\subsubsection{FaceForensics++}
The FF++ dataset \cite{ff++} contains video footage of faces that were faked using four common face faking methods: Deepfakes (DF) \cite{deepfake}, Face2Face (F2F) \cite{thies2016face2face}, FaceSwap (FS) \cite{FaceSwap}, and Nulltextures (NT) \cite{xception}. The original video footage was obtained from YouTube, including 1000 real videos and 4000 fake videos. In order to simulate different qualities, the FF++ dataset is available in both high quality (HQ) and low-quality versions (i.e., c23 and c40).

\subsubsection{Celeb-DF(V2)}
The Celeb-DF(V2) dataset \cite{celebdf} consists of 590 real videos and 5639 fake videos, all of which are 30 seconds long. The original videos come from YouTube public videos and cover a wide distribution of gender, age and ethnicity.
Celeb-DF(V2) uses an improved DeepFake algorithm to generate high-resolution faces, which employs a codec with more layers and increased dimensionality. Also, a color conversion algorithm is introduced to address issues such as inconsistent facial colors, and the quality of the generated video is improved by adding training data and post-processing.

\subsubsection{DFDC}
The DFDC \cite{dfdc} dataset is currently the largest publicly available dataset in the field, containing real videos from 3,426 paid actors. The dataset generates more than 100,000 fake videos through a variety of faking methods, including DeepFakes methods, GAN methods, and non-deep learning methods.

\subsubsection{WildDeepfake}
This database  contains 7,314 facial action sequences extracted from 707 Deepfake videos, all of which are rich and diverse from the web. These facial action sequences are extracted to make the visual effects more realistic and more in line with real-life scenarios.

\subsubsection{DFF}
A total of 30,000 real images and 90,000 fake images were generated from the original IMDB-WIKI \cite{imdb} dataset using the Stable Diffusion Inpainting and InsightFace toolbox methods respectively.

\begin{figure}[t] 
\centering 
\caption{Visualization of typical samples from five datasets, i.e., FF++ \cite{ff++}, Celeb-DF (v2) \cite{celebdf}, DFF \cite{dff}, WDF \cite{wildfake}, and DFDC \cite{dfdc}. } 
\includegraphics[width=0.48\textwidth]{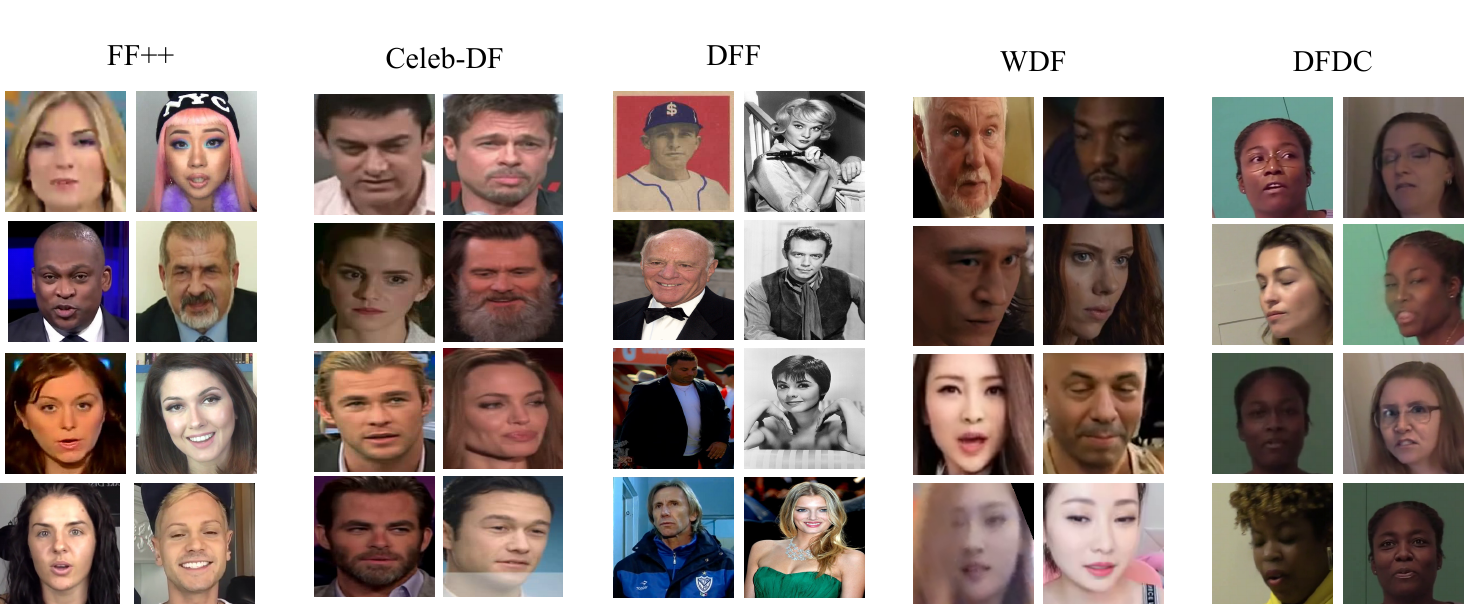}

\label{difface} 
\end{figure}


\vspace{0.2em}
\noindent\textbf{Protocols.} Although some existing studies have proposed different forgery methods for single-dataset training.No pilot study is available for training on multiple datasets with real-world diverse forgery patterns and large-scale characteristics . Besides the perspective of training, only a single forgery test domain is usually used in the evaluation of algorithm performance, which leads to biased comparisons of state-of-the-art methods. To tackle the above-mentioned issues, we provide a novel data arrangement and training/testing strategy to benchmark the fair evaluations. Specifically, $5$ datasets (each dataset is regarded as an individual domain), i.e., FF++ \cite{ff++}, WDF \cite{wildfake}, Celeb \cite{celebdf}, DFDC \cite{dfdc}, and DFF \cite{dff} are merged into a large set $D$, which can be further divided into training sets $\left\{ D_{\text{FF++}} ,D_{\text{WDF}},D_{\text{Celeb}},D_{\text{DFF}}\right\}$ and test set $\left\{D_{\text{DFDC}},D_{\text{i}} \right\}$. $i= \begin{Bmatrix}\text{FF++},\text{WDF},\text{Celeb},\text{DFF}\end{Bmatrix}$ which denotes the subsets removed from the training set for the testing set. Specifically, in consideration of costly training time, the large-scale DFDC \cite{dfdc} is only used for testing. We randomly select $n\leq3$ subsets of the data for training. $n=1$ for the traditional single-domain training protocol while $n=3$ denotes the newly established multi-domain protocols:
$\left\{ D_{\text{FF++}} \cup D_{\text{WDF}} \cup D_{\text{Celeb}} \right\}, \left\{ D_{\text{FF++}} \cup D_{\text{WDF}} \cup D_{\text{DFF}} \right\}, \\ \left\{ D_{\text{FF++}} \cup D_{\text{FF++}} \cup D_{\text{DFF}} \right\}$ and $\left\{ D_{\text{WDF}} \cup D_{\text{Celeb}} \cup D_{\text{DFF}} \right\}$ for training, respectively. More details can be found in the supplementary material.




\noindent\textbf{Evaluation metrics.}  Three common metrics, i.e., Accuracy (ACC (\%)), Area Under ROC Curve (AUC (\%)), and Equal Error Rate (EER (\%)) are adopted. EER is defined as the error rate when false acceptance rate (FAR) is equal to false rejection rate (FRR), and can be expressed as $EER=\frac{FRR+FAR}{2}$. However, in the realm of evaluating face forgery detection, we are more concerned with keeping the FAR at a relatively low level to ensure that it will not be easy to authenticate a forged face. For this purpose, we introduce the a priori probability of positive examples when calculating the EER. Since the impact on the system of a positive sample misclassified as a negative example is much greater than the impact of a negative sample as a positive example. To counteract this effect, we introduced the $P_{real}$ parameter.
In addition, we found that the original EER did not take into account the effect of testing on multiple domains. Calculating the EER directly on each dataset and then averaging the values may be affected by extreme values, and we judged performance against multiple domains by taking the maximum of instead of simply averaging them. This ensures that our evaluation is more accurate and robust.

\begin{equation}
\begin{gathered}
M^{i}={P^{i}_{real}*FRR^{i}+(1-P^{i}_{real})*FAR^{i}}\\
M_{ERR}=Max\left\{ M^{1},M^{2},\cdots M^{N} \right\}
\end{gathered}
\end{equation}

Here $P_{real}$ is the prior probability of the real samples. $M^{i}$ is the prior probability EER of the $i$-th target domain. $N$ is the number of target test domains.

\section{Experiments}
In this section, we evaluate the performance of our proposed method on FaceForensics++ (FF++) \cite{ff++}, Celeb-DF(V2) \cite{celebdf}, DFDC \cite{dfdc} and WildDeepfake \cite{wildfake} datasets on both traditional protocols as well as the proposed benchmark.

\subsection{Implementation Details}
We use \text{ViT-B/16} \cite{clip} as the backbone model. We uses RetinaFace \cite{retinaface}  to detect facial areas and scaled the face image to $224\times224$ with a patch size of $16$. We trained the model using the Adam optimizer with the learning rate set to 3e-6. The batch size during training was 32, and 40 training epochs were performed.Following official setting \cite{ff++} , we extracted 100 frames from each video as validation set and test set. During the training process, only random flipping was used for data augmentation.During the dataset merging phase, multiple datasets are randomly shuffled and then consolidated into a new dataset. All code is implemented using the PyTorch framework.	

\begin{table*}[]
\caption{Cross-Manipulation Evaluation: ACC (\%) and AUC (\%) for Multi-Source Training and Testing.}
    %
    \label{tradition1}
    \centering
    \resizebox{0.8\textwidth}{!}{\begin{tabular}{ccccccccc}
      \hline
      \multirow{2}{*}{Method} & \multicolumn{2}{c}{GID-DF (C23)} & \multicolumn{2}{c}{GID-DF (C40)} & \multicolumn{2}{c}{GID-F2F (C23)} & \multicolumn{2}{c}{GID-F2F (C40)} \\ \cline{2-9} 
                               & ACC(\%)             & AUC(\%)            & ACC(\%)             & AUC(\%)            & ACC(\%)             & AUC(\%)             & ACC(\%)             & AUC(\%)             \\ \hline
      EfficientNet \cite{tan2019efficientnet} $\left( PMLR \,2019\right)$           & 82.40           & 91.11          & 67.60           & 75.30          & 63.32           & 80.10           & 61.41           & 67.40           \\
      ForensicTransfer \cite{cozzolino2018forensictransfer} $\left( Arxiv \,2018\right)$       & 72.01           & -              & 68.20           & -              & 64.50           & -               & 55.00           & -               \\
      Multi-task \cite{nguyen2019multi}  $\left( BTAS \,2019\right)$            & 70.30           & -              & 66.76           & -              & 58.74           & -               & 56.50           & -               \\
    F$^3$-Net \cite{qian2020thinking} $\left( ECCV \,2020\right)$              & 83.57           & 94.95          & 77.50           & 85.77          & 61.07           & 81.20           & 64.64           & 73.70           \\
      MLGD \cite{li2018learning}    $\left( AAAI \,2018\right)$                & 84.21           & 91.82          & 67.15           & 73.12          & 63.46           & 77.10           & 58.12           & 61.70           \\
      LTW \cite{sun2021domain}         $ \left( AAAI \,2021\right)$           & 85.60           & 92.70          & 69.15           & 75.60          & 65.60           & 80.20           & 65.70           & 72.40           \\
      DCL \cite{sun2022dual}       $\left( AAAI \,2022\right) $             & 87.70           & 94.90          & 75.90           & 83.82          & 68.40           & 82.93           & 67.85           & 75.07           \\
      M2TR \cite{wang2022m2tr}     $\left( ICML  \,2022\right)$               & 81.07           & 94.91          & 74.29           & 84.85          & 55.71           & 76.99           & 66.43           & 71.70           \\

      Implicit \cite{implicit}    $\left( CVPR \,2023\right)$         & 88.21           & 95.03          & 76.90           & 84.55         & 69.36          & 84.37         & 67.99           & 74.80         \\ 


      
 \textbf{GM-DF (Ours)}                     & \textbf{91.34}           & \textbf{96.62}    & \textbf{78.13}     & \textbf{85.19}    & \textbf{72.32}     & \textbf{86.30}    &\textbf{69.02}     & \textbf{75.51}      \\ \hline
    \end{tabular}}
\end{table*}

\subsection{Preliminary Multi-datasets Investigation}
\noindent\textbf{Single-dataset:} Each dataset is trained separately to be evaluated on different test sets, and the cross-dataset performance of the model is tested on different dataset datasets using the model trained from scratch.

\noindent\textbf{Direct-Merged datasets:} After directly merging multiple face forgery datasets, we employ a straightforward strategy of training a baseline fake detector using a generalized classification loss. This aims to verify whether directly merging the datasets is expected to improve the existing forgery detection models across datasets.

To investigate the feasibility of co-training from multiple datasets to improve the cross-dataset performance, we use multiple datasets from different sources for training. We also add the recently proposed which use stable diffusion generated face data for exploration, we believe is more in line with the real-world nature of forgery methods.  Table \ref{compare2} shows the following essential findings:

1) \textit{Directly combining datasets for training does not increase accuracy.} We first train the baseline detector on a single dataset (e.g., FF++ \cite{ff++} or Celeb \cite{celebdf} as well as DFF \cite{dff} ) and evaluate this trained baseline on two different datasets (eg. WDF \cite{wildfake} and DFDC \cite{dfdc} ). As shown in Table \ref{compare2} the baseline performs well only on its original training dataset (e.g., 75.49\% AUC to 67.19\%). Its detection accuracy drops severely when the baseline detector combines the dataset Celeb-DF(V2) (with an accuracy of only 53.12\%). This is mainly due to the fact that the single dataset detection model over-fits the common features of its training dataset, but does not take into account the variations in the characteristics of the source-to-target dataset. The same accuracy degradation problem can be observed on other baseline detectors such as REECE \cite{recce} and recent sota model \cite{implicit,ucf} .

2) \textit{Pre-trained model performs well within domain and modelling knowledge fading in unseen samples.} We fine-tuned the baseline model on both individual and multiple datasets, and subsequently compared the outcomes.We observed that the model trained on either the FF++ or Celeb dataset performed better when trained individually, whereas its performance deteriorated when trained jointly on both datasets. Table \ref{compare2} shows that the model is pretrained on FF++ and Celeb-DF(V2), but the detection accuracy of DFDC \cite{dfdc} is still  poor because the model has been fine-tuned to Celeb \cite{celebdf}, forgetting what was learned from the previous pretraining dataset and the differences in joint training at different resolutions.

3) \textit{Training stable diffusion and GAN face forgery data together may increase these differences and learning difficulties.} For example, after fusing DFF \cite{dff}(the dataset generated by the diffusion model) in the WDF \cite{wildfake} test the results of the Xception \cite{xception} model dropped the most from 67.19\% to 58.82\% , which has a higher demand on the detector.Both recent sota model UCF \cite{ucf} and Implicit \cite{implicit} exhibit similar characteristics in the experimental results, leading to an unavoidable performance decline. Specifically, UCF \cite{ucf} achieves 97.10\% on Celeb training and testing, but rapidly drops to 73.71\% when tested on the DFF dataset.	
 
4) \textit{The effectiveness of the proposed method.} Table \ref{compare2} shows that the average cross- and intra-domain detection performance of the proposed method exceeds the baseline models, proving its flexibility and validity. Our model demonstrates an average AUC improvement of 8.87\% over UCF \cite{ucf} in cross-domain scenarios and 6.53\% in within-domain scenarios, underscoring the superiority of the two-stage learning approach.	
 

\begin{table}[t]
\caption{Cross-domain comparisons of generalization based on AUC (\%). We train the model on the HQ dataset of FF++ \cite{ff++} and evaluate it on Celeb-DF(V2) \cite{celebdf} and DFDC\cite{dfdc} .}\label{tab2:crossdomain}
\centering 
\tabcolsep=1cm
  \scalebox{0.8}{
\begin{tabular}{l|ccc}
\hline
Method                          & Celeb       & DFDC         \\ \hline
EN-B4 ~\cite{efb4}   $\left( PMLR \,2019\right)$                 & 66.24          & 66.81          \\
F$^3$-Net ~\cite{f3net}  $\left( ECCV \,2020\right)$    & 71.21          & 72.88          \\
Xception ~\cite{xception}  $\left( ICCV \,2021\right) $           & 66.91          & 69.93         \\
MAT(EN-B4) ~\cite{mat}    $ \left( CVPR \,2021\right)$             & 76.65          & 67.34         \\
Face X-ray ~\cite{spsl}      $\left( CVPR \,2021\right) $          & 74.20          & 70.00          \\
RFM ~\cite{rfm}         $\left( CVPR \,2021\right) $             & 67.64          & 68.01       \\
SRM ~\cite{srm}   $\left( CVPR \,2021\right) $                      & 79.40          & 79.70         \\
Local-relation ~\cite{localrelation}   $\left( AAAI \,2021\right)$  & 78.26          & 76.53          \\
LTW ~\cite{ltw}  $\left( AAAI \,2021\right)$                     & 77.14          & 74.58        
   \\ 
RECCE ~\cite{recce}         $\left( CVPR \,2022\right)$            & 77.39          & 76.75         \\

Impliciry ~\cite{implicit}   $\left( ICCV \,2023\right)$                   & 82.04          & \text{-}         
   \\ 

SFGD ~\cite{SFDG}    $\left( CVPR \,2023\right)$                   & 75.83          & 73.64         
   \\ 
\textbf{GM-DF (Ours)}                         & \textbf{83.16} & \textbf{77.23} \\ 
\hline

\end{tabular}}
%
 
\end{table}

\subsection{Results on the Proposed Protocols}
To further assess the real-world performance of our method, we conducted experiments on the proposed multi-datasets deepfake detection benchmark (in Sec. \ref{protocls}). We compared our method with commonly used forgery detection networks such as MesoNet \cite{mesonet} and Xception \cite{xception}, as well as some recent state-of-the-art (SOTA) methods. RECCE \cite{recce},UCF \cite{ucf} and Implicit \cite{implicit}  was tested under default settings. As shown in Table \ref{protocol}, according to the results under evaluation metrics $M_{EER}$ \& AUC, the proposed method outperforms other methods, showcasing its effectiveness. An intriguing finding emerged: Despite achieving excellent performance on individual datasets, some existing SOTA methods experience drastic performance drop under multi-datasets protocols.

\begin{table}[]
\caption{Natural language descriptions of the
real and fake face used to train the model. BLIP Generate indicates that the BLIP \cite{blip} model generates descriptive information.}
\label{prompt}
\resizebox{0.47\textwidth}{!}{%
\begin{tabular}{ccc}
\cline{1-3}
\multicolumn{1}{c}{Prompt} & \multicolumn{1}{c}{Real Prompt} & \multicolumn{1}{c}{Fake Prompt}  \\ \cline{1-3}
\multicolumn{1}{c}{P1} & \multicolumn{1}{c}{A photo of real face} & \multicolumn{1}{c}{A photo of fake face}  \\
\multicolumn{1}{c}{P2} & \multicolumn{1}{c}{This is a photo of real} & \multicolumn{1}{c}{This is a photo of fake}  \\
\multicolumn{1}{c}{P3} & \multicolumn{1}{c}{\{BLIP Generate \}  A photo of real face} & \multicolumn{1}{c}{\{BLIP Generate \} A photo of real face} \\
\multicolumn{1}{c}{P4} & \multicolumn{1}{c}{Real} & \multicolumn{1}{c}{Fake}  \\
\multicolumn{1}{c}{P5} & \multicolumn{1}{c}{This is how a real face looks like} & \multicolumn{1}{c}{This is how a fake face looks like} \\ 
\multicolumn{1}{c}{P6} & \multicolumn{1}{c}{This photo contains real face} & \multicolumn{1}{c}{This photo contains fake face}\\ 
\multicolumn{1}{c}{P7} & \multicolumn{1}{c}{Real face is in this photo} & \multicolumn{1}{c}{Fake face is in this photo}\\ 
\cline{1-3}
\end{tabular}%
}

\end{table}

\subsection{Results on Traditional Protocols}
We also conducted experiments on the commonly used benchmarks \cite{sun2022dual} to prove the effectiveness of our methodology in classic single dataset multi-source operation settings. For this experimental setup, we selected one class of manipulated forged videos from FF++ \cite{ff++} as the unseen manipulation sample, while utilizing the remaining three classes as the training set. The experiment results are in Table \ref{tradition1}. All four assessment settings show our model achieved better detection results. Our Model achieves 3.13\% improvement in ACC on GID-DF (C23) compared to  recent sota model Implicit \cite{implicit}, proving that our model can better adapt to forgery methods in multiple source domains and learn the common and characteristic features of each forgery method.

To gain a more detailed understanding of cross-domain performance, we employed a single data training approach. Specifically, our model was trained on FF++ (C23) \cite{ff++} and tested on DFDC \cite{dfdc} and Celeb-DF(V2) \cite{celebdf}. Comparative results with state-of-the-art methods are presented in Table \ref{tab2:crossdomain}. In cross-dataset comparisons, our model demonstrates excellent performance. In internal tests, when compared to the recent REECE \cite{recce}, our model exhibits a notable improvement of  4.17\%  and 3.59\% compare to SFGD \cite{SFDG} in AUC.This indicates that our model outperforms traditional cross-forgery patterns and other state-of-the-art models in terms of generalization, showcasing the transfer capabilities of GM-DF models. This underscores the effectiveness of natural language supervision in generating more generalizable representations, particularly in the context of cross-dataset training data.

\begin{figure*}[h] 
\centering 
\label{Datavis}
\includegraphics[width=1\textwidth]{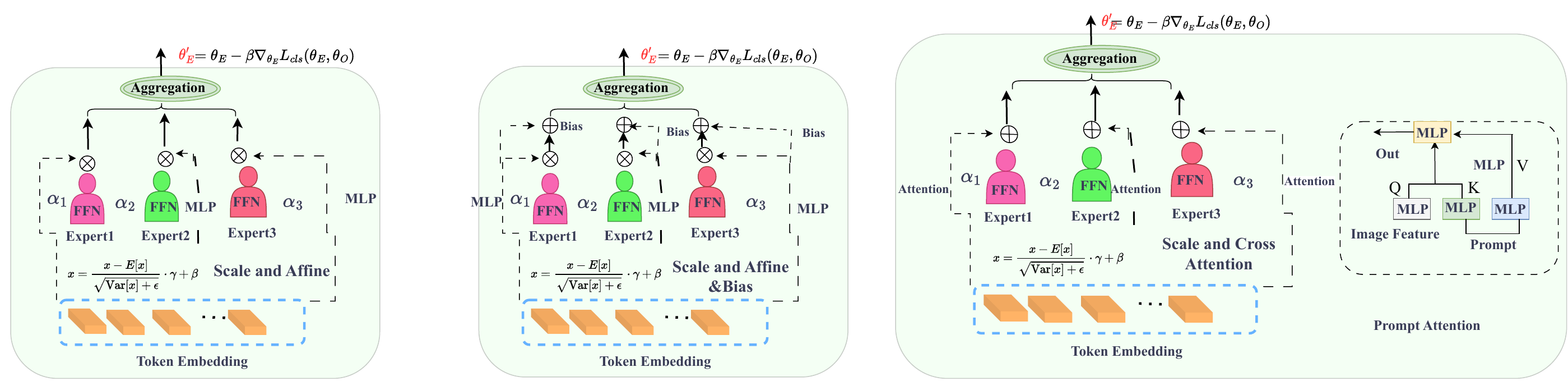}
\caption{Architectures of three adaptation strategies for theDataset Information Layer, including Affine (left), Affine\&Bias (middle), and Cross Attention (right).} 
\label{diffnorm} 
\end{figure*}

\begin{table}[]
\caption{Impact of different text prompts (described
in Table \ref{prompt}).}
\label{prompt_compare}
\resizebox{0.47\textwidth}{!}{%
\begin{tabular}{lcccccc}
\cline{1-7}
\multicolumn{1}{l}{\multirow{2}{*}{Method}} & \multicolumn{2}{c}{FF + Celeb-\textgreater{}WDF} & \multicolumn{2}{c}{FF +DFF-\textgreater{}WDF} & \multicolumn{2}{c}{Celeb +DFF-\textgreater{}WDF}  \\ \cline{2-7}
\multicolumn{1}{l}{} & \multicolumn{1}{c}{AUC(\%)} & \multicolumn{1}{c}{EER(\%)} & \multicolumn{1}{c}{AUC(\%)} & \multicolumn{1}{c}{EER(\%)} & \multicolumn{1}{c}{AUC(\%)} & \multicolumn{1}{c}{EER(\%)} \\ \cline{1-7}
\multicolumn{1}{l}{P1} & \multicolumn{1}{c}{63.08} & \multicolumn{1}{c}{\textbf{29.23}} & \multicolumn{1}{c}{\textbf{76.09}} & \multicolumn{1}{c}{\textbf{30.95}} & \multicolumn{1}{c}{\textbf{78.09}} & \multicolumn{1}{c}{\textbf{31.27}} \\ 
\multicolumn{1}{l}{P2} & \multicolumn{1}{c}{61.30} & \multicolumn{1}{c}{29.85} & \multicolumn{1}{c}{75.88} & \multicolumn{1}{c}{34.56} & \multicolumn{1}{c}{77.92} & \multicolumn{1}{c}{31.92}  \\ 
\multicolumn{1}{l}{P3} & \multicolumn{1}{c}{61.37} & \multicolumn{1}{c}{31.81} & \multicolumn{1}{c}{69.93} & \multicolumn{1}{c}{32.03} & \multicolumn{1}{c}{74.25} & \multicolumn{1}{c}{34.87} \\ 
\multicolumn{1}{l}{P4} & \multicolumn{1}{c}{62.11} & \multicolumn{1}{c}{30.67} & \multicolumn{1}{c}{70.25} & \multicolumn{1}{c}{31.87} & \multicolumn{1}{c}{72.22} & \multicolumn{1}{c}{33.89} \\ 
\multicolumn{1}{l}{P5} & \multicolumn{1}{c}{\textbf{}\textbf{63.33}} & \multicolumn{1}{c}{30.35} & \multicolumn{1}{c}{74.40} & \multicolumn{1}{c}{33.30} & \multicolumn{1}{c}{77.18} & \multicolumn{1}{c}{31.46}  \\ 
\multicolumn{1}{l}{P6} & \multicolumn{1}{c}{\textbf{}60.43} & \multicolumn{1}{c}{35.17} & \multicolumn{1}{c}{72.20} & \multicolumn{1}{c}{34.51} & \multicolumn{1}{c}{76.54} & \multicolumn{1}{c}{34.46}  \\ 
\multicolumn{1}{l}{P7} & \multicolumn{1}{c}{\textbf{}61.10} & \multicolumn{1}{c}{33.29} & \multicolumn{1}{c}{72.43} & \multicolumn{1}{c}{34.69} & \multicolumn{1}{c}{78.12} & \multicolumn{1}{c}{33.15}  \\ 
\cline{1-7}
\end{tabular}%
}

\end{table}

\begin{table}[t]
\caption{Ablation of each component on the protocol of FF++\& Celeb\& DFF to WDF.}
\label{abalation}
\begin{tabular}{ccccc|cc}
\hline
ID & Baseline & DA & MIM & Meta-MoE & AUC & ACC \\ \hline
$\Rmnum{1}$ & \checkmark &  &  &  & 73.45 & 71.07 \\
$\Rmnum{2}$ & \checkmark & \checkmark &  &  & 74.36 & 71.65 \\
$\Rmnum{3}$ & \checkmark & &\checkmark  &  & 75.11 & 73.33 \\
$\Rmnum{4}$ & \checkmark & &  &\checkmark   & 77.21 & 74.18\\
$\Rmnum{5}$ & \checkmark & \checkmark & \checkmark &  & 75.71 & 73.42 \\
$\Rmnum{6}$ & \checkmark & \checkmark &  & \checkmark & 75.12 & 74.39 \\

GM-DF (Ours) & \checkmark & \checkmark & \checkmark & \checkmark & \textbf{79.70} & \textbf{75.13} \\ \hline
\end{tabular}%
\end{table}

\begin{figure}[h] 
\centering 
\label{Datavis}
\includegraphics[width=0.3\textwidth]{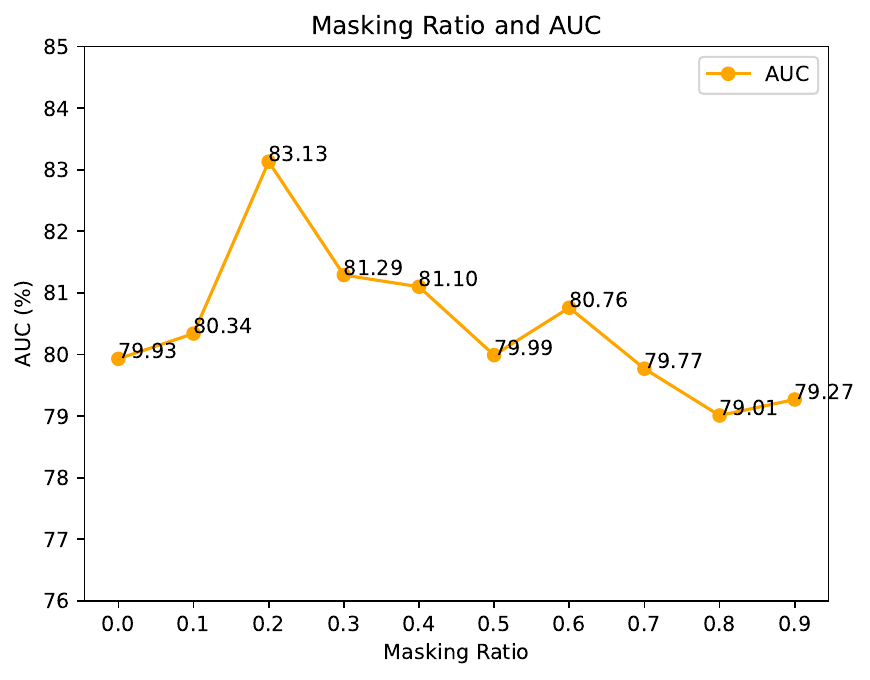}
\caption{Quantitative analyses of masking strategy. The AUC (\%) scores of cross-dataset evaluation
on Celeb-DF are reported.} 
\label{mask} 
\end{figure}


\begin{figure*}[t]
    \setlength{\abovecaptionskip}{0.cm}
    \begin{center}
        \includegraphics[width=1\linewidth]{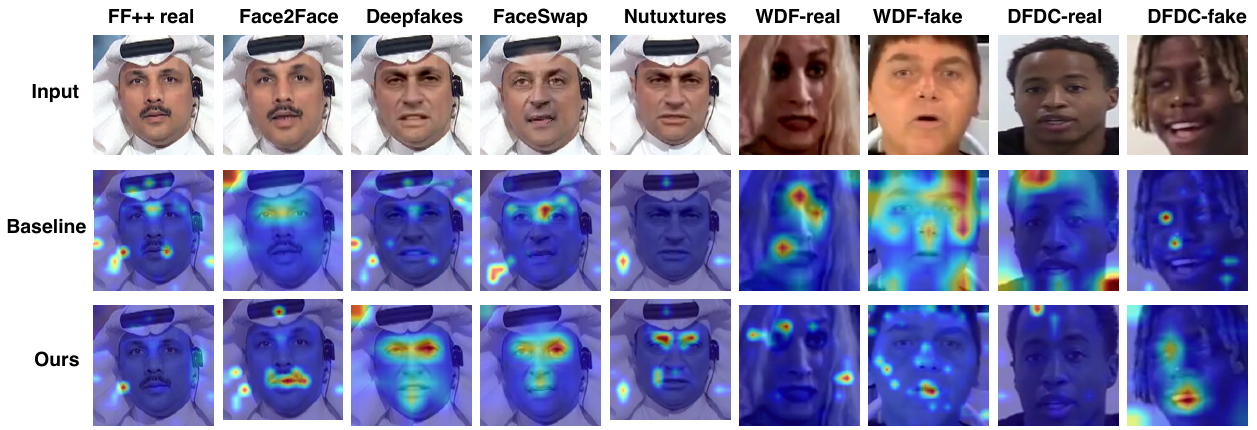}
    \end{center}
    \caption{
The model's attention is illustrated through a heatmap, where darker colors signify increased focus in that specific region. The first column represents the input image, the second column depicts the outcome obtained by directly fusing the data using fine-tuning with the CLIP \cite{clip}, and the third column showcases the results from our model. 
}
    \label{features}
\end{figure*} 

\subsection{Ablation Study}
In this subsection, we train on Celebdf \cite{celebdf} and FF++ \cite{ff++} datasets and test on WDF \cite{wildfake} to test our proposed module and essential parameter settings.

\vspace{0.2em}
\noindent\textbf{Effectiveness of different text prompts.} To validate the effect 
of different prompts on experimental performance, we 
 introduced new templates into the  prompts 
group. In Table \ref{prompt},  shows the specific language descriptions 
of the real and fake face categories.
we scrutinize the impact of distinct text prompts on the model. Notably, varied texts exhibit commendable performance across diverse datasets, with marginal differences. This substantiates the notion that text can effectively manifest dynamicized parameters in real-world contexts, thereby affirming our concept of instating dynamic affine transformations tailored to each dataset. An intriguing discovery emerges when utilizing BLIP \cite{blip} to generate images with detailed descriptive information alongside the original combination of category images. Surprisingly, performance experiences relative degradation, potentially attributed to interference induced by category-independent prompts.

\vspace{0.2em}
\noindent\textbf{Impacts of various ViT backbone initialization.} To extend our observations on the impact of initialization on the multi-datasets training, we tuned the model using different CLIP pre-training weights and showed a comparison of their performance in Table \ref{prompt_compare}. Specifically, we fine-tuned the weighting using two architectures, Resnet and Vit,\ a)  Resnet50 backbone; \ b)  Resnet101 backbone; \ c)  ViT backbone with a patch size of 16; \ d) ViT backbone with a patch size of 32; and \ e)  ViT backbone with a patch size of 14. It can be seen that ViT pre-training initialization yields better multi-dataset training generalization compared to other initialization methods
Compared to other initialization methods, the Transformer initialization achieves better multi-datasets training generality due to its powerful representation extraction capability, which provides a better image-text alignment basis and detailed feature extraction capability for all image alignment experiments.


\vspace{0.2em}
\noindent\textbf{Effectiveness of DA loss.}
In the \ref{abalation} first and second rows compared to the baseline, the DA loss achieved an improvement of approximately { 0.91\%}, demonstrating the need for the alignment of the distributions of the two feature datasets through higher-order statistical features. We can observe a consistent improvement in performance when using the DA loss function, which demonstrates the advantage of dataset alignment with the visual-linguistic pre-training model.

\vspace{0.2em}
\noindent\textbf{Effectiveness of MIM loss.}
Comparing the first and third rows, it can be seen that the addition of the reconstruction module improves the AUC by 1.66\%  over the original model, which indicates that the reconstructed features can effectively enhance the ability of fine-grained information extraction on the forged face.

\begin{figure}[t] 
\centering 
\label{Datavis}
\includegraphics[width=0.34\textwidth]{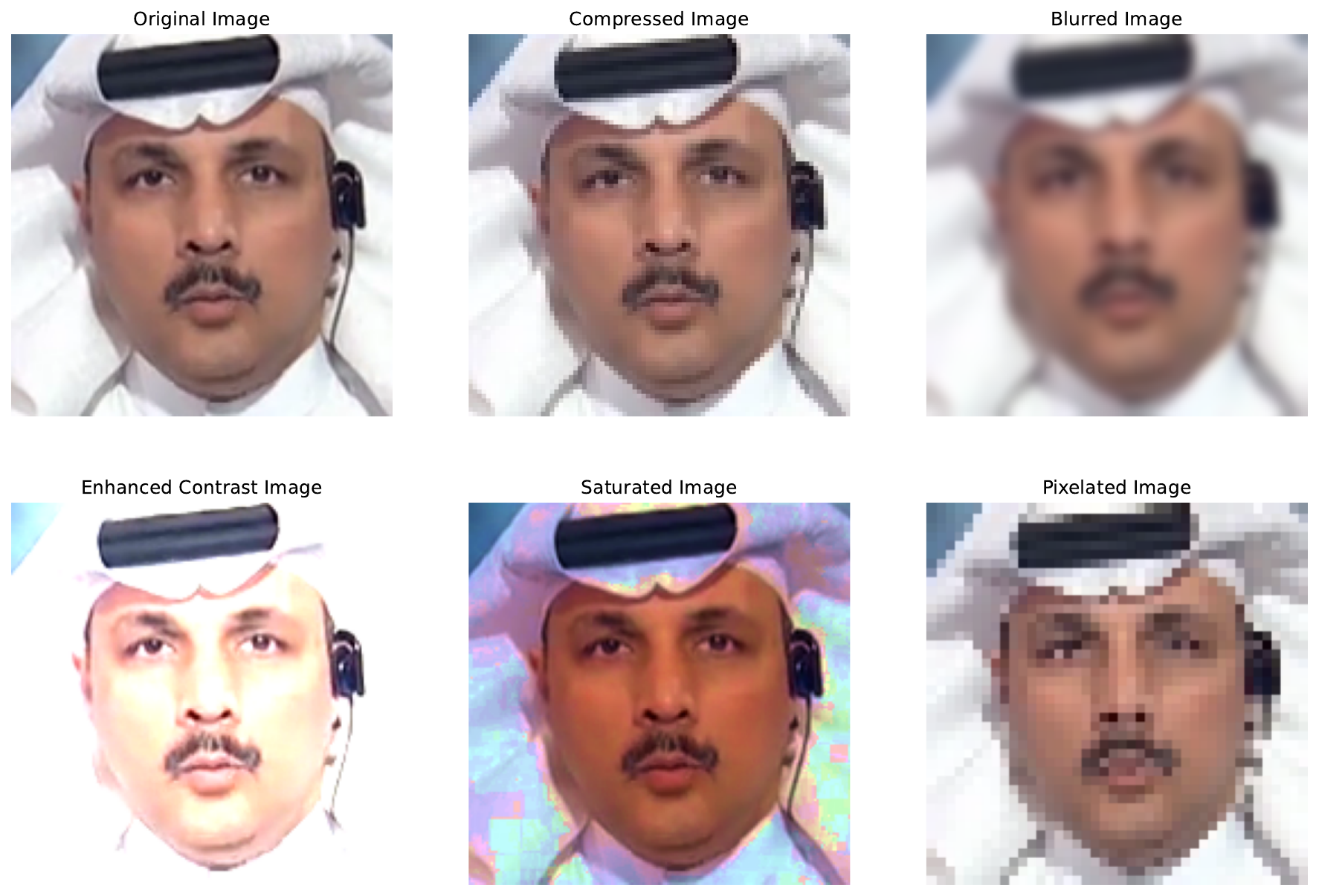}
\caption{Examples of images with different quality-degradation methods.Image
Compression, Gaussian Blur, Enhanced Contrast dithering, Satureted dithering, and pixelization, respectively. } 

\label{degration} 
\end{figure}

\begin{table}[h]
\centering
\caption{Results of different domain adaptive strategies when trained on Celeb-DF (v2) \cite{celebdf} \& DFF \cite{dff} and tested on WDF \cite{wildfake} dataset.}
\resizebox{0.35\textwidth}{!}{%
\begin{tabular}{ccc}
\hline
\multirow{2}{*}{Method} & \multicolumn{2}{c}{Celeb-DF (v2) \& DFF} \\ \cline{2-3} 
 & ACC(\%) & AUC(\%) \\ \hline
\textbf{Affine} & \textbf{65.21} & \textbf{67.32} \\
Affine \& Bias & 63.24 & 66.13 \\
Cross Attention & 64.18 & 66.87 \\ \hline
\end{tabular}%
}

\label{affine}
\end{table}

\vspace{0.2em}
\noindent\textbf{Effectiveness of Meta-MoE.}
To quantify the importance of Meta-Moe module, we compare our text-based supervisory signals with meta learning and without two stage learning.It can be seen from the fourth line that meta-MoE plays an important role in performance improvement (from 73.45\% to 77.21\%), which is mainly caused by learning the characteristic features of the domain.The mask-supervision method exhibits better generalization, suggesting that mask supervision alone can restrain
overfitting to the training data.Moreover, unlike the only text backbone we improves steadily with more fine-grained supervision,, which further confirms the scalability and versatility of multidataset learning .

\vspace{0.2em}
\noindent\textbf{Analysis of masking ratio.}
The quantitative results of the cross-dataset evaluation are shown in Figure \ref{mask}. We observe that the minimum and randomized masking strategy achieves optimality under medium masking rates. Their performance is severely degraded as the masking rate is greater than 80\%. The random masking strategies work best at 20\% maskingrate.This indicates that some important face edges may be corrupted using the random masking strategy.

\subsection{Visualization and Discussion}

\noindent\textbf{Discussion about the Dataset Information Layer}.
To address the challenge of feature adaptation to different dataset domains, as illustrated in Figure \ref{difface}, we 
also investigate three different domain adaptation strategies (i.e., Affine, Affine\&Bias, and Cross Attention) for the Dataset Information Layer.

\textbf{1) Affine.} The domain-specific knowledge of each domain can be realized by linearly mapping the respective prompt feature multiplication to the intermediate feature layer. This part of the linear mapping is realized through a single MLP.We also visualized the Adaptive layer features, which suggests that the differences of different data sets are effectively learned.

\textbf{2) Affine\&Bias.}
Here, we adjust the LayerNorm's parameters via learning by both affines and offsets. The vanilla LayerNorm assumes that the samples are all from the same distribution,which the data might come from different domains. Therefore, the parameters in LayerNorm should be not the same in different domains. The LayerNorm based on Affine\&Bias learning can be formulated as follows:
\begin{equation}
\text{LayerNorm}(x) = \frac{x - E[x]}{\sqrt{\text{Var}[x] + \epsilon}} \cdot \left(  \gamma*\gamma_{d} \right) +\left(  \beta_{d}  \right).
\end{equation}
Domain-specific parameters $\gamma*\gamma_{d}$ and  $\beta+ \beta_{d} $  can adaptively change the intermediate representation conditions and domain indicators capture distinctive characteristics.

\textbf{3) Cross Attention.}
Based on \cite{Detr,towards}, we use Cross Attention to aggregate text features and raw features with a cross attention block with jump connections at the beginning of each encoder-decoder stage.
First, we partitioned the domain cues into n independent in-domain cue embeddings that have the same shape, which partially acts as a reference set for cross-attention, with the images providing the associated information. Next, a series of attention operations are performed between the query vectors generated for each image and the key-value vectors generated for the domain cues. Finally, the results of the attention operations are added to the data point embeddings after projection by a zero-initialized linear layer.To validate the effectiveness of our model in Figure 4 we use to visualize the ROC curves, the data were trained in FF++c23 and Celeb and tested on various datasets .

The results of these three domain adaptive strategies when trained on Celeb-DF (v2) \cite{celebdf} \& DFF \cite{dff} and tested on WDF \cite{wildfake} are shown in Table \ref{affine}. We can find that the Affine strategy is simple yet effective, and achieves better cross-domain performance than other two alternatives. Besides, we also find that the performance of Cross Attention strategy seems satisfactory, and one possible future direction is how to efficiently combine Affine with Cross Attention to boost generalization capacity.

\vspace{0.2em}
\noindent\textbf{Analysis of robustness against distortions}.
Considering the prevalence of image processing on the web, we investigate the performance under several distortions proposed by \cite{multitask,recce}, namely image compression, Gaussian blurring, contrast dithering, saturation dithering and pixelization.The quality-degraded
images using different degradation methods are shown in
Fig 8. The results are shown in Table \ref{roubst}. We can see that our model is more robust to the listed ingressions than the existing methods. Both our method and previous methods are generally robust to compression, contrast and saturation. However, in scenarios blur and pixelate, the performance \cite{xception,rfm,f3net,multitask,recce} are still much lower than the proposed method, indicating the robustness of the proposed method.

\begin{table}[]
\caption{Robustness evaluation in terms of AUC (\%) on WildDeepfake (WDF) dataset. }
\resizebox{0.5\textwidth}{!}{%
\begin{tabular}{ccccccc}
\hline
Method & Compress & Blur & Contrast & Saturate & Pixelate & Avg \\ \hline
Multi-task \cite{multitask} $\left( BTAS \,2019\right)$& 89.64 & 80.98 & 89.30 & 90.37 & 79.44 & 85.95 \\
F$^3$-Net \cite{f3net} $\left( ECCV \,2020\right)$ & 86.71 & 78.99 & 86.53 & 87.67 & 73.23 & 82.63 \\
Xception \cite{xception} $\left( ICCV \,2021\right)$ & 86.01 & 78.29 & 81.90 & 84.96 & 66.24 & 79.48 \\
RFM \cite{rfm} $\left( ECCV \,2021\right)$  & 83.74 & 75.34 & 79.77 & 82.59 & 71.25 & 78.54 \\
Add-Net \cite{add}  $\left( AAAI \,2021\right)$& 83.34 & 79.66 & 84.46 & 85.13 & 64.33 & 79.38 \\
REECE \cite{recce}  $\left( CVPR \,2022\right)$& 89.65 & 87.29 & 91.19 & 91.74 & 83.88 & 88.75 \\



\textbf{\textbf{GM-DF (Ours)}} & \textbf{90.32} & \textbf{89.43} & \textbf{92.56} & \textbf{92.31} & \textbf{84.95} & \textbf{89.91} \\ \hline
\end{tabular}%
}

\label{roubst}
\end{table}

\vspace{0.2em}
\noindent\textbf{Visualization}.  We employed a joint training approach using three datasets FF++ \cite{ff++}, Celeb-DF (V2) \cite{celebdf}, and DFF \cite{dff}. Subsequently, we conducted visual analyses on individual in-domain datasets as well as various cross-domain datasets. From Figure \ref{features}, it can be observed that directly merging datasets often leads the model to lose effective focus in challenging scenarios, such as WDF \cite{wildfake}, where attention shifts to background regions. In contrast, our proposed multi-domain fusion model consistently concentrates on facial regions and successfully detects manipulated faces.

\section{Conclusion}
In this paper, we investigate the generalization capacity of deepfake detectors when trained on multi-dataset scenarios and propose a novel benchmark for multi-scenario training. We design a Generalized Multi-Scenario Deepfake Detection (GM-DF) framework to learn of both specific and common features across datasets. By utilizing generic text representations to learn the relationships across different datasets, we propose a novel meta-learning strategy to capture the relational information among datasets. Besides, GM-DF employs contrastive learning on image-text pairs to capture common dataset characteristics and utilizes self-supervised mask relation learning to mask out partial correlations between regions during training. Extensive experiments demonstrate the superior generalization of our method. In the future, we plan to explore techniques for localizing counterfeit regions and enhancing generalization by leveraging multimodal large language models.


\bibliographystyle{IEEEtran}
\bibliography{IEEEabrv,reference}

\end{document}